\documentclass[letterpaper]{article} 
\usepackage[draft]{aaai2026}  
\usepackage{times}  
\usepackage{helvet}  
\usepackage{courier}  
\usepackage[hyphens]{url}  
\usepackage{graphicx} 
\usepackage{natbib}  
\usepackage{caption} 
\usepackage{algorithm}
\usepackage{algorithmic}
\usepackage{amsmath}
\usepackage{amsthm}  
\usepackage{booktabs}
\usepackage{multirow}
\usepackage{amssymb}
\usepackage{amsfonts}
\usepackage{subcaption}
\usepackage{times}
\usepackage{helvet}
\usepackage{courier}
\usepackage{xcolor}
\theoremstyle{plain}  

\theoremstyle{remark}  
\newtheorem*{proof-sketch}{Proof Sketch}

\newcounter{checksubsection}
\newcounter{checkitem}[checksubsection]

\usepackage{newfloat}
\usepackage{listings}
\DeclareCaptionStyle{ruled}{labelfont=normalfont,labelsep=colon,strut=off} 
\floatstyle{ruled}
\newfloat{listing}{tb}{lst}{}
\floatname{listing}{Listing}
\title{Beyond a Single Sign: BiTE for Branch-Preserving Binary Threshold Expansion}
\author{
    Siddharth Gupta,
    Mikail Yayla,
    Akash Kumar
}

\affiliations{
    Ruhr University Bochum, Germany\\
    siddharth.gupta@rub.de,
    mikail.yayla@rub.de,
    akash.kumar@rub.de
}

\begin{document}

\maketitle

\begin{abstract}
Binary neural networks (BNNs) reduce model storage and replace many arithmetic
operations with efficient bitwise computations, but the single-sign activation
limits their representational capacity. Each preactivation direction is
compared with one threshold and reduced to one of two values, discarding
variation within each sign region. We propose \textbf{BiTE}
(Branch-Preserving Binary Threshold Expansion), which evaluates each feature
direction against $K$ ordered thresholds and retains the resulting binary
responses as separate inputs to the following learned operator. BiTE can
distinguish up to $K+1$ ordered activation states while allowing the following
operator to learn a separate transformation for each threshold response. It
also enables a controllable trade-off between the number of independently
constructed feature directions and the number of ordered responses extracted
from each direction. On BinaryViT, BiTE improves mean final-epoch Top-1
accuracy from $74.60\pm0.30\%$ to $75.83\pm0.15\%$ on CIFAR-10 and from
$42.80\pm0.46\%$ to $44.07\pm0.32\%$ on CIFAR-100, with nearly unchanged
trainable parameter counts. A matched uniform-collapse control reaches
$75.33\pm0.15\%$ on CIFAR-10, supporting the value of retaining the threshold
responses separately relative to the evaluated scalar collapse. Binary VGG-16
experiments further show a favorable observed parameter--accuracy trade-off,
including a configuration that achieves $0.64$ percentage points higher mean
accuracy with $21.8\%$ fewer parameters than conventional width scaling.
Overall, these results support branch-preserving threshold expansion as an
effective way to improve binary representations and the observed
parameter--accuracy trade-off.
\end{abstract}

\section{Introduction}
\label{sec:intro}

\begin{figure*}[t]
    \centering
    \includegraphics[width=\textwidth]{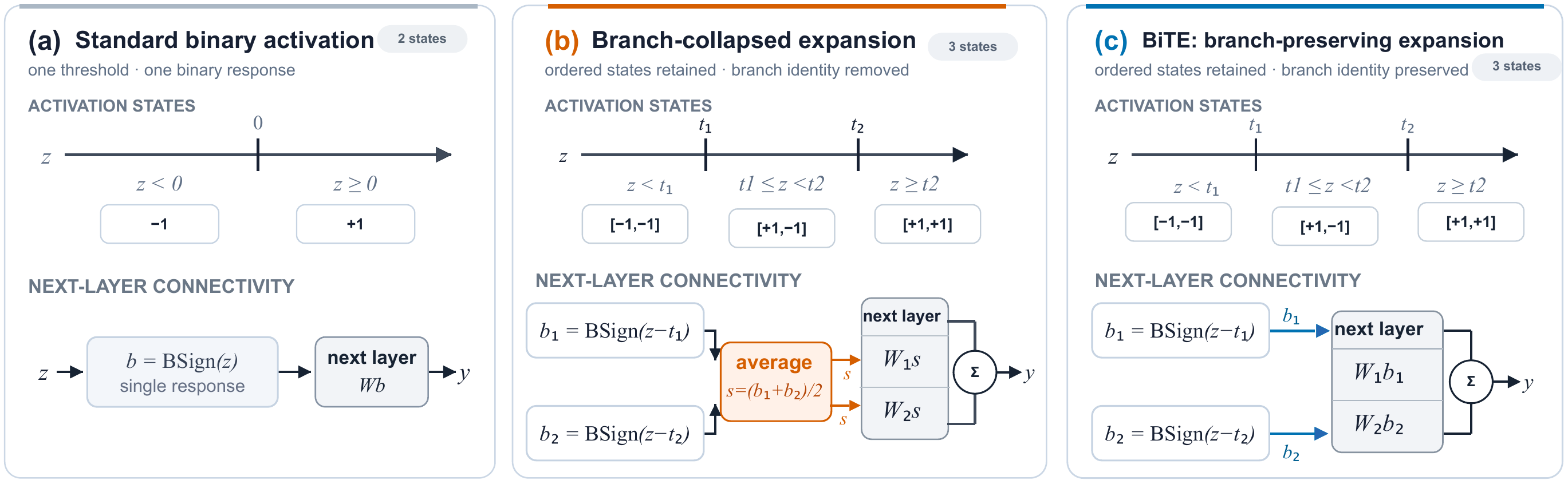}
   \caption{\textbf{From a single sign to branch-preserving threshold
expansion.}
(a) A conventional sign activation produces one binary response per feature
direction.
(b) Two ordered thresholds produce three activation states, but uniform
collapse forces both threshold crossings to share the same downstream effect.
(c) BiTE retains the threshold responses as separate inputs, enabling
threshold-specific downstream transformations.}
    \label{fig:bite_overview}
\end{figure*}

Binary neural networks (BNNs) represent weights and activations with one bit,
enabling bitwise implementations of the dominant inner products while reducing
weight and activation storage~\cite{hubara2016binarized,rastegari2016xnor}.
This efficiency is accompanied by substantial representational loss. In
particular, the standard sign activation partitions each feature direction at
one boundary and maps all values on either side to the same binary response.
Activation magnitude and variation within each sign region are therefore
unavailable to the following layer.

Recent BNNs reduce the accuracy loss through scaling factors, real-valued
shortcuts, improved optimization, adaptive thresholds, and richer binary
operators
~\cite{rastegari2016xnor,liu2018bi,qin2020forward,xu2021recu,
ding2022ienet,liu2020reactnet,zhang2022dynamic,lee2023insta}.
Most nevertheless expose only one binary response at each feature coordinate.
Multiple-binarization and low-bit quantization methods introduce additional
binary components or activation states, but typically combine them through
scalar reconstruction, residual approximation, or aggregation
~\cite{lin2017towards,ghasemzadeh2018rebnet,zhuang2019structured,
zhou2016dorefa,choi2018pact,esser2019learned}. These approaches motivate a complementary question: can several ordered
binary responses be extracted from one learned feature direction while
remaining separately accessible to the next learned operator?
Figure~\ref{fig:bite_overview} contrasts the conventional single sign
interface, a scalar collapsed representation, and the proposed
branch preserving formulation.

We answer this question with \textbf{BiTE}, a branch-preserving binary
threshold-expansion mechanism. For each preactivation direction $z_g$, BiTE
learns $K$ ordered thresholds and produces

\[
    b_{g,k}
    =
    \operatorname{BSign}(z_g-t_{g,k}),
    \qquad
    k=1,\ldots,K.
\]

These responses define up to $K+1$ ordered activation states. Rather than
combining them into a scalar before the next transformation, BiTE retains
each direction--threshold pair as a distinct input. The consuming operator
can therefore assign different learned effects to different threshold
crossings of the same feature direction.

Branch preservation also induces a capacity-allocation trade-off. In the
fixed-width convolutional construction, an interface of width $C$ uses
$C/K$ preactivation directions and extracts $K$ responses from each. BiTE
therefore reallocates capacity from constructing additional directions toward
representing more ordered states along each retained direction. In
BinaryViT, the responses temporarily expand the MLP representation and are
then mapped back to the original width by a learned reduction. Thresholds
are optimized jointly with the network and stored as constants at inference.

\noindent Our main contributions are:
\begin{itemize}
    \item \textbf{Branch-preserving threshold expansion.}
    We introduce BiTE, which extracts multiple ordered binary responses from
    each learned feature direction and retains them as separate inputs to the
    next operator, enabling threshold-specific downstream transformations.

    \item \textbf{Architecture-compatible capacity allocation.}
    We develop a fixed-width convolutional formulation that trades
    independently constructed directions for multiple ordered responses per
    direction, and a BinaryViT formulation that preserves these responses
    through temporary expansion and learned reduction.

    \item \textbf{Mechanism controls and cross-architecture evaluation.}
    We compare BiTE with single-sign, uniform-collapse, and fixed-threshold
    controls in BinaryViT, study parameter--accuracy trade-offs in binary
    VGG-16, and evaluate transfer to an IE-Net-based binary ResNet-18.
\end{itemize}

\section{Related Work}
\label{sec:related}

Prior work improves binary representations through stronger backbones,
adaptive thresholding, multiple binary components, and feature expansion.
We organize these directions according to how they relate to BiTE.

\paragraph{Binary neural networks.}
Early BNNs introduced binary weights and later extended binarization to both
weights and activations
~\cite{courbariaux2015binaryconnect,hubara2016binarized}.
XNOR-Net added scaling factors for binary convolutions
~\cite{rastegari2016xnor}, while later methods improved information flow,
optimization, and gradient estimation through real-valued shortcuts and
specialized binary operators
~\cite{liu2018bi,qin2020forward,xu2021recu,ding2022ienet,gupta2025ab}.
These methods strengthen the binary backbone but do not explicitly retain
multiple ordered responses from the same preactivation direction.

\paragraph{Adaptive binary activations and thresholds.}
ReActNet replaces the fixed sign activation with RSign, which learns a
channel-wise shift, and reshapes the post-binary distribution with
RPReLU~\cite{liu2020reactnet}. Dynamic BNN and INSTA-BNN make thresholds
input-dependent using feature statistics
~\cite{zhang2022dynamic,lee2023insta}, whereas AdaBin learns the center and
distance of the two-valued binary set~\cite{tu2022adabin}. LAB replaces the
pointwise sign function with a learned fine-grained binarization kernel
~\cite{falkena2023lab}, while DeSign applies spatially patterned thresholds
following a dithering construction, allowing adjacent positions to use
different binarization boundaries~\cite{monroy2024designed}. These methods
modify how one binary response is produced at each coordinate. BiTE instead
evaluates multiple ordered thresholds for the same feature direction and
position and preserves all resulting responses for the next transformation.

\paragraph{Multiple binary bases and multilevel quantization.}
ABC-Net represents weights and activations using linear combinations of
binary bases~\cite{lin2017towards}, ReBNet uses residual binarization to
create multiple activation levels~\cite{ghasemzadeh2018rebnet}, and
Group-Net aggregates several binary transformations
~\cite{zhuang2019structured}. Similarly, multibit methods such as
DoReFa-Net, PACT, and LSQ partition the activation range into multiple
intervals and encode the selected interval as a scalar
~\cite{zhou2016dorefa,choi2018pact,esser2019learned}. These approaches
reconstruct a value or transformation from multiple components. BiTE instead
retains the ordered binary responses as separate input coordinates, allowing
the following operator to learn distinct effects for each
direction--threshold pair.

\paragraph{Representation expansion and capacity allocation.}
ES-BNN is closely related to BiTE from a capacity-allocation perspective. It
replicates binary feature channels, reduces the number of channels produced
directly by the binary convolution to preserve its parameter and operation
counts, and relies on channel-wise normalization and residual paths to
diversify the initially repeated features~\cite{shi2025expanding}. Its
transformer instantiation averages expanded channels before attention to
recover the original attention width. BiTE does not construct its expanded
interface through direct replication: each retained direction produces
multiple ordered threshold-indexed responses that remain independently
transformable. In the fixed-width convolutional instantiation, BiTE
reallocates interface capacity from independently constructed directions to
ordered responses per direction. The BinaryViT instantiation instead uses
temporary threshold expansion followed by a learned branch-preserving
reduction.

\paragraph{Binary vision transformers.}
Binarizing vision transformers introduces additional difficulties in
self-attention, token interaction, and MLP representations. The BinaryViT
architecture of Le and Li incorporates CNN-inspired pooling, residual, and
pyramidal components to improve binary feature representation
~\cite{le2023binaryvit}. A separate BinaryViT formulation uses gradient
regularization and activation shifting to stabilize fully binarized ViTs
~\cite{xiao2024binaryvit}. Bi-ViT addresses attention distortion through
gradient reactivation and ranking-aware distillation~\cite{li2024bi};
BVT-IMA modifies binary attention using learned information tables
~\cite{wang2024bvt}; and BHViT combines a binarization-friendly hybrid
architecture with dedicated MLP and attention modules
~\cite{gao2025bhvit}. These methods primarily redesign transformer
architectures, attention computation, or training objectives. BiTE instead
modifies the MLP activation interface by preserving multiple ordered responses
from each intermediate feature as separately transformable inputs to the
following projection.

\section{BiTE: Branch-Preserving Threshold Expansion}
\label{sec:method}

A conventional binary activation compares each feature direction with one
threshold and produces a single binary response. BiTE instead uses $K$
ordered thresholds, yielding $K$ binary responses that can distinguish up to
$K+1$ ordered activation regions. These threshold-indexed responses, which
we call branches, remain separate so that the following learned operator can
assign distinct effects to different threshold crossings. Here, a branch
denotes a threshold-indexed feature coordinate rather than a separate
subnetwork or physical module. Figure~\ref{fig:bite_overview} contrasts
single-sign activation, uniformly collapsed threshold expansion, and the
proposed branch-preserving formulation.

\subsection{Ordered Threshold Responses}
\label{sec:threshold_encoding}

Let $\mathbf{Z}=\{z_g(p)\}$ denote the real-valued tensor presented to the
threshold comparisons, where $g\in\{1,\ldots,G\}$ indexes a feature direction
and $p$ denotes a spatial or token position. A conventional binary activation
produces $b_g(p)=\operatorname{BSign}(z_g(p))$, where

\begin{equation}
    \operatorname{BSign}(v)
    =
    \begin{cases}
        +1, & v\geq 0,\\
        -1, & v<0.
    \end{cases}
\label{eq:bsign}
\end{equation}

BiTE assigns $K$ learnable raw thresholds
$\boldsymbol{\rho}_g=[\rho_{g,1},\ldots,\rho_{g,K}]$ to each feature
direction. The thresholds are sorted separately for each feature direction
during every forward pass:

\begin{equation}
    [t_{g,1},\ldots,t_{g,K}]
    =
    \operatorname{sort}(\boldsymbol{\rho}_g),
    \qquad
    t_{g,1}\leq\cdots\leq t_{g,K}.
\label{eq:ordered_thresholds}
\end{equation}

Each ordered threshold produces a separate one-bit response,

\begin{equation}
    b_{g,k}(p)
    =
    \operatorname{BSign}\!\left(z_g(p)-t_{g,k}\right),
    \qquad
    k=1,\ldots,K.
\label{eq:bite_response}
\end{equation}

For the principal setting $K=2$,

\begin{equation}
    [b_{g,1}(p),b_{g,2}(p)]
    =
    \begin{cases}
        [-1,-1], & z_g(p)<t_{g,1},\\
        [+1,-1], & t_{g,1}\leq z_g(p)<t_{g,2},\\
        [+1,+1], & z_g(p)\geq t_{g,2}.
    \end{cases}
\label{eq:bite_three_states}
\end{equation}

Because the ordered responses share the same preactivation, $K$ distinct
thresholds yield $K+1$ reachable monotonic response patterns rather than all
$2^K$ possible binary combinations. Across $G$ feature directions, BiTE
concatenates the resulting $GK$ one-bit responses and exposes them separately
to the following learned operator.

\subsection{Why Branch Preservation Matters}
\label{sec:branch_preservation}

Although $K$ ordered thresholds distinguish up to $K+1$ activation regions,
the following operator can assign distinct effects to their crossings only if
the corresponding responses remain separate. For one output location of the
immediately following affine operator, let $\mathcal{P}$ denote its spatial
or token input positions, and let
$\mathbf{w}_{g,k}(p)\in\mathbb{R}^{O}$ connect response $b_{g,k}(p)$ to
$O$ output coordinates. Then

\begin{equation}
    \mathbf{y}
    =
    \mathbf{c}
    +
    \sum_{p\in\mathcal{P}}
    \sum_{g=1}^{G}
    \sum_{k=1}^{K}
    \mathbf{w}_{g,k}(p)b_{g,k}(p).
\label{eq:bite_output}
\end{equation}

As $z_g(p)$ increases across threshold $t_{g,k}$, the output changes by
$2\mathbf{w}_{g,k}(p)$. Hence, the effects of the $K$ crossings are

\begin{equation}
    \mathbf{E}^{\mathrm{BiTE}}_g(p)
    =
    2
    \begin{bmatrix}
        \mathbf{w}_{g,1}(p) & \cdots & \mathbf{w}_{g,K}(p)
    \end{bmatrix}
    \in\mathbb{R}^{O\times K}.
\label{eq:bite_effect_matrix}
\end{equation}

Its columns are learned separately, so the matrix can attain rank up to
$\min(O,K)$.

By contrast, consider a scalar encoding
\[
    q_g(p)
    =
    \sum_{k=1}^{K}\alpha_{g,k}b_{g,k}(p),
\]
where
$\boldsymbol{\alpha}_g=
[\alpha_{g,1},\ldots,\alpha_{g,K}]^{\mathsf T}$,
followed by $\mathbf{v}_g(p)\in\mathbb{R}^{O}$. Its threshold-effect matrix is

\begin{equation}
    \mathbf{E}^{\mathrm{coll}}_g(p)
    =
    2\mathbf{v}_g(p)\boldsymbol{\alpha}_g^{\mathsf T},
    \qquad
    \operatorname{rank}
    \left(\mathbf{E}^{\mathrm{coll}}_g(p)\right)
    \leq 1.
\label{eq:collapsed_rank}
\end{equation}

The scalar code may still retain up to $K+1$ ordered states, but its threshold
crossings affect the immediate output through collinear vectors that differ
only by scalar coefficients. Branch preservation removes this local rank-one
restriction by keeping the threshold responses separately transformable.

\paragraph{Matched uniform-collapse control.}
To isolate branch preservation relative to uniform scalar collapse, we average
the $K$ responses and repeat the resulting scalar:

\begin{equation}
    s_g(p)
    =
    \frac{1}{K}\sum_{k=1}^{K}b_{g,k}(p),
    \qquad
    \mathbf{b}^{\mathrm{coll}}_g(p)
    =
    \operatorname{Repeat}_{K}\!\left(s_g(p)\right).
\label{eq:collapsed_state}
\end{equation}

This control preserves the representation width and the same ordered scalar
code, with up to $K+1$ states. For $K=2$ with distinct thresholds, these
states are $-1$, $0$, and $+1$. However, because every repeated coordinate
receives the same value, all threshold crossings affect the immediate output
along the same direction. Comparing this control with BiTE therefore tests the
value of retaining separately transformable threshold responses relative to
uniform collapse, not relative to every possible learned scalar encoding.

\subsection{Architecture Instantiations}
\label{sec:architecture_instantiations}

\paragraph{Fixed-width convolutional instantiation.}
Let $C$ denote the one-bit interface width presented to the following
convolution, with $C$ divisible by $K$. A conventional binary layer
constructs $C$ feature directions and extracts one binary response from each.
BiTE instead constructs $G=C/K$ directions and extracts $K$ ordered
responses from each:

\begin{equation}
\begin{aligned}
    \text{conventional BNN:}\quad
    &C\times 1=C,\\
    \text{BiTE:}\quad
    &\frac{C}{K}\times K=C.
\end{aligned}
\label{eq:fixed_width_allocation}
\end{equation}

The following convolution therefore receives $C$ one-bit channels in both
cases, with the $K$ responses of each retained direction occupying separate
input channels. BiTE thus reallocates the fixed interface width from
independently constructed feature directions to multiple ordered responses
per retained direction.

If the producer has input width $C_{\mathrm{in}}$ and $\kappa$ kernel
positions, its binary-weight count changes from
$\kappa C_{\mathrm{in}}C$ to $\kappa C_{\mathrm{in}}C/K$. The following
convolution retains the same input width, while BiTE introduces
$GK=C$ learnable threshold scalars. 

\paragraph{BinaryViT instantiation.}
Let $\mathbf{Z}\in\mathbb{R}^{B\times N\times D}$ denote the output of
the first quantized projection in a BinaryViT MLP block, where $D$ is the
intermediate width. For the $k$th ordered threshold vector
$\mathbf{t}_k\in\mathbb{R}^{D}$, BiTE forms

\begin{equation}
    \mathbf{B}_k
    =
    \operatorname{BSign}\!\left(\mathbf{Z}-\mathbf{t}_k\right)
    \in\{-1,+1\}^{B\times N\times D},
    \qquad
    k=1,\ldots,K,
\label{eq:vit_threshold_groups}
\end{equation}

where the thresholds are broadcast across batches and tokens. These binary
groups are concatenated with the bounded base feature
$\mathbf{Z}_{\mathrm{base}}=\operatorname{hardtanh}(\mathbf{Z})$:

\begin{equation}
    \mathbf{A}_{\mathrm{ViT}}
    =
    \operatorname{Concat}\!\left(
        \mathbf{Z}_{\mathrm{base}},
        \mathbf{B}_1,\ldots,\mathbf{B}_K
    \right)
    \in\mathbb{R}^{B\times N\times D(K+1)}.
\label{eq:vit_expansion}
\end{equation}

A learned reduction maps the expanded representation back to width $D$.
Partitioning its weights by input group gives

\begin{equation}
    \mathbf{U}
    =
    \mathbf{Z}_{\mathrm{base}}\mathbf{W}_{\mathrm{base}}
    +
    \sum_{k=1}^{K}\mathbf{B}_k\mathbf{W}_k
    \in\mathbb{R}^{B\times N\times D}.
\label{eq:vit_branch_reduction}
\end{equation}

Each threshold group has its own learned transformation $\mathbf{W}_k$,
preserving threshold-specific effects through the reduction. The reduced
tensor then passes through the block normalization and RPReLU before the
original BinaryViT output projection.

The reduction uses the same quantized-linear weight operator as the
BinaryViT MLP, with one-bit weights in the reported experiments. Its input
activation precision is configured at 32 bits, so
$\mathbf{Z}_{\mathrm{base}}$ remains real-valued and the groups
$\mathbf{B}_k$ enter the reduction directly rather than being binarized
again.

For the matched collapsed-state control, the $K$ threshold groups are
replaced by $K$ copies of their average. The bounded base path, expanded
width, reduction layer, and subsequent operations remain identical. Thus, the
two variants differ only in whether the reduction receives separate threshold
responses or their uniformly collapsed representation.

\subsection{Training and Inference}
\label{sec:training_inference}

For $K\geq2$, each feature direction is initialized with uniformly spaced
thresholds in $[-a,a]$. Unless otherwise stated, we use $K=2$ and $a=1$,
giving initial thresholds of $-1$ and $+1$.
During training, the raw thresholds are optimized jointly with the remaining
network parameters and sorted independently for each feature direction in every
forward pass, as defined in Eq.~\eqref{eq:ordered_thresholds}. Away from exact
ties, sorting acts as a permutation, allowing automatic differentiation to route
each ordered-threshold gradient to its corresponding raw threshold parameter.

The forward pass uses the hard response in Eq.~\eqref{eq:bite_response}. Let
$\psi$ denote the surrogate activation of the underlying binary backbone.
Backpropagation applies its straight-through estimator to both the preactivation
and threshold:

\begin{equation}
\begin{aligned}
    \frac{\partial b_{g,k}(p)}{\partial z_g(p)}
    &\approx
    \psi'\!\left(z_g(p)-t_{g,k}\right),\\
    \frac{\partial b_{g,k}(p)}{\partial t_{g,k}}
    &\approx
    -\psi'\!\left(z_g(p)-t_{g,k}\right).
\end{aligned}
\label{eq:threshold_ste}
\end{equation}

BiTE therefore reuses the backbone's gradient surrogate without introducing a
threshold-specific training objective. A module with $G$ feature directions and
$K$ thresholds contains $GK$ real-valued threshold scalars. At inference, the
final ordered thresholds are stored as constants, so runtime sorting is
unnecessary; each preactivation is computed once and compared with the $K$
stored thresholds.

\section{Experiments}
\label{sec:experiments}

We evaluate whether BiTE improves binary-network accuracy across datasets and
architectures, whether preserving threshold responses is beneficial relative
to uniform scalar collapse, and how its fixed-width capacity allocation
compares with conventional width scaling.

\subsection{Experimental Setup}
\label{sec:experimental_setup}

\paragraph{Datasets, architectures, and configuration selection.}
We evaluate BinaryViT on CIFAR-10 and CIFAR-100
~\cite{krizhevsky2009learning}, an IE-Net-based binary ResNet-18 and
binary VGG-16 on CIFAR-10, and BinaryViT on ImageNet
~\cite{deng2009imagenet}. DeSign is included as an independently trained
convolutional reference. Using one exploratory seed-0 CIFAR-10 run, we
evaluated five layer placements with $K=2$ and additionally tested $K=3$
in layers 0--6. We selected $K=2$ in layers 0--6. This run is included in
the three-run CIFAR-10 summary, whereas CIFAR-100 did not inform
configuration selection. Complete exploratory results are provided in the
supplementary material.

\paragraph{Training and reporting.}
BinaryViT models on CIFAR are trained for 300 epochs using AdamW and a cosine
learning-rate schedule. Thresholds use the same learning rate as the remaining
parameters and zero weight decay. We report final-epoch Top-1 accuracy on the
CIFAR test set as the sample mean and sample standard deviation over
seeds 0, 1, and 2.
IE-Net and IE-Net with BiTE use three matched runs and report the highest
test accuracy observed during training. DeSign uses three independently
trained runs and is included as a contextual reference rather than a matched
ablation.

Binary VGG-16 models are trained for 500 epochs over three independent runs.
We report both final-epoch accuracy and the highest accuracy recorded on the
CIFAR test set. Because no independent test score is available after checkpoint selection, final-epoch accuracy is the primary measure in the parameter--accuracy
analysis.
The ImageNet study uses one run per configuration, 100 training epochs, and a
global batch size of 512. We report the checkpoint with the highest validation
Top-1 accuracy, with Top-5 measured at the same checkpoint. Complete
optimization, augmentation, scheduling, and batch-size settings are provided
in the supplementary material.

\paragraph{Parameter accounting.}
Parameter counts are measured from the instantiated models and include learned
thresholds and architecture-specific reduction layers. Efficiency comparisons
concern trainable parameters only; operation count, activation storage, energy,
and latency are not evaluated.

\subsection{Accuracy Across Datasets}
\label{sec:main_results}

\begin{table}[t]
\centering
\small
\setlength{\tabcolsep}{5.0pt}
\begin{tabular}{llc}
\toprule
\textbf{Dataset} &
\textbf{Method} &
\textbf{Top-1 (\%)} \\
\midrule
\multirow{2}{*}{CIFAR-10}
    & BinaryViT
    & $74.60 \pm 0.30$ \\
    & BinaryViT + BiTE
    & $\mathbf{75.83 \pm 0.15}$ \\
\midrule
\multirow{2}{*}{CIFAR-100}
    & BinaryViT
    & $42.80 \pm 0.46$ \\
    & BinaryViT + BiTE
    & $\mathbf{44.07 \pm 0.32}$ \\
\bottomrule
\end{tabular}
\caption{\textbf{BinaryViT results on CIFAR-10 and CIFAR-100.}
Values are final-epoch Top-1 accuracy on the CIFAR test set, reported
as sample mean $\pm$ sample standard deviation over seeds 0, 1, and 2.}
\label{tab:main_results}
\end{table}

BiTE increases mean final-epoch accuracy by $1.23$ percentage points on
CIFAR-10 and $1.27$ points on CIFAR-100. The CIFAR-10 result includes the
seed-0 run used for configuration selection, whereas CIFAR-100 did not inform
that selection.

On CIFAR-10, the instantiated baseline and BiTE models contain $21.615$M and
$21.600$M trainable parameters, respectively. Thus, the observed improvement
does not result from an increase in total parameter count. These totals cover
the complete models rather than only the thresholds and reduction layer.

\subsection{Secondary CNN Evaluation}
\label{sec:cnn_transfer}

We also evaluate BiTE using an IE-Net-based binary ResNet-18 on CIFAR-10.
IE-Net and IE-Net with BiTE use three matched runs. DeSign is included as
an independently trained reference rather than a matched ablation.
\begin{table}[t]
\centering
\small
\setlength{\tabcolsep}{5.0pt}
\begin{tabular}{lc}
\toprule
\textbf{Method} &
\textbf{Top-1 (\%)} \\
\midrule
DeSign
    & $89.04 \pm 0.18$ \\
IE-Net
    & $91.89 \pm 0.05$ \\
IE-Net + BiTE
    & $\mathbf{92.06 \pm 0.13}$ \\
\bottomrule
\end{tabular}
\caption{\textbf{Secondary evaluation on binary ResNet-18 and CIFAR-10.}
Values report the highest test accuracy observed during training as sample
mean $\pm$ sample standard deviation over three runs. IE-Net and IE-Net with
BiTE use matched runs, whereas DeSign is independently trained.}
\label{tab:cnn_transfer}
\end{table}
Adding BiTE to stages 1--3 increases the mean accuracy of IE-Net from
$91.89\pm0.05\%$ to $92.06\pm0.13\%$. The paired changes are
$+0.29$, $+0.01$, and $+0.20$ percentage points. BiTE adds $2{,}688$
trainable parameters, corresponding to approximately $0.024\%$ of the
baseline model. The positive change in all three matched runs provides supporting evidence
that BiTE transfers to a convolutional architecture.

DeSign reaches $89.04\pm0.18\%$ over three independent runs. IE-Net with
BiTE has a $3.02$ point higher mean, but this is not a controlled comparison
because the methods use different binary backbones and training procedures.
The matched IE-Net comparison therefore provides the direct evidence for the
incremental contribution of BiTE.

\begin{figure*}[t]
    \centering
    \includegraphics[width=0.92\textwidth]
    {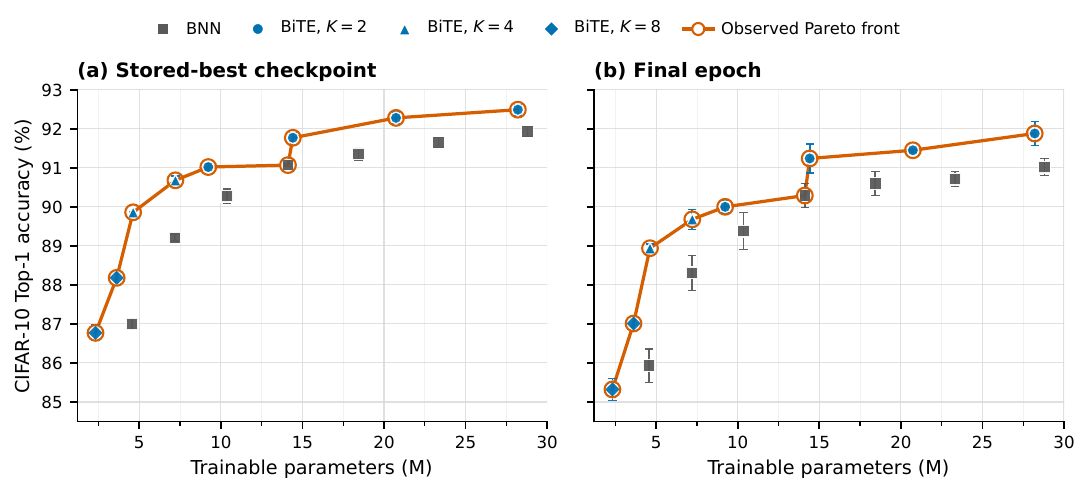}
    \caption{\textbf{Parameter--accuracy trade-off on VGG-16/CIFAR-10.}
(a) Highest test-set accuracy observed during training and
(b) final-epoch test accuracy. Points report the mean over three runs, with
error bars showing the sample standard deviation. Nondominance is determined
from the mean parameter--accuracy coordinates among the evaluated
configurations; connecting lines are visual guides only.}
    \label{fig:parameter_accuracy}
\end{figure*}

\subsection{Mechanism Controls}
\label{sec:mechanism_ablation}

We compare the single-sign baseline with uniform scalar collapse,
fixed-threshold BiTE, and learnable BiTE. All multi-threshold variants use
$K=2$, modify layers 0--6, and use the same expanded reduction topology.
\begin{table}[t]
\centering
\small
\setlength{\tabcolsep}{3.2pt}
\begin{tabular}{lccc}
\toprule
\textbf{Variant} &
\textbf{Thresholds} &
\textbf{Responses} &
\textbf{Top-1 (\%)} \\
\midrule
Single-sign baseline
    & One at $0$
    & Single
    & $74.60 \pm 0.30$ \\
Collapsed-state control
    & Learned
    & Collapsed
    & $75.33 \pm 0.15$ \\
Fixed-threshold BiTE
    & Fixed
    & Preserved
    & $75.63 \pm 0.25$ \\
BiTE
    & Learned
    & Preserved
    & $\mathbf{75.83 \pm 0.15}$ \\
\bottomrule
\end{tabular}
\caption{\textbf{Mechanism controls on BinaryViT/CIFAR-10.}
Values are final-epoch Top-1 accuracy on the CIFAR test set, reported
as sample mean $\pm$ sample standard deviation over seeds 0, 1, and 2.}
\label{tab:mechanism}
\end{table}
The collapsed-state control exceeds the single-sign baseline by $0.73$
points. Because this variant also introduces the expansion-and-reduction topology,
the baseline-to-control difference does not isolate the additional ordered
state; a duplicated-sign control with the same topology is not evaluated.

BiTE exceeds the collapsed-state control by $0.50$ points. Both variants use
learned thresholds, the same expanded dimensionality, and the same reduction
topology. They differ only in whether the threshold responses remain separate
or are averaged and repeated before reduction. The result is therefore
consistent with a benefit from preserving threshold responses as separately
transformable features relative to the evaluated uniform scalar collapse. It
does not cover learned scalar encodings such as
$q_g=\sum_k\alpha_{g,k}b_{g,k}$ and is not a comparison against all
multilevel scalar activations.

Fixed-threshold BiTE exceeds the baseline by $1.03$ points without optimizing
the threshold positions. Learning the thresholds yields a further observed
difference of $0.20$ points. Given three runs and the overlapping variation,
this difference alone does not establish a reliable benefit from threshold
learning.

\subsection{Parameter--Accuracy Trade-off}
\label{sec:parameter_accuracy}

We vary the width scale $\rho$ of binary VGG-16 and the threshold count $K$
of BiTE to compare accuracy across parameter budgets.

At approximately $4.6$M parameters, BiTE with $K=4$ reaches
$88.94\pm0.11\%$, compared with $85.93\pm0.43\%$ for the conventional
BNN. This corresponds to a $3.01$ point higher mean accuracy at a nearly
matched parameter count.
BiTE with $K=2$ reaches $91.24\pm0.37\%$ using $14.410$M parameters,
whereas the $18.436$M-parameter conventional BNN reaches
$90.60\pm0.31\%$. At these evaluated operating points, BiTE has a
$0.64$-point higher mean final-epoch accuracy with approximately $21.8\%$
fewer parameters.

The threshold count controls the fixed-width allocation. At
$\rho=1.5$, increasing $K$ from $2$ to $4$ and $8$ reduces the parameter
count from $9.228$M to $4.622$M and $2.318$M, respectively, while mean
final-epoch accuracy changes from $90.00\%$ to $88.94\%$ and $85.32\%$.
Larger $K$ therefore allocates more threshold responses to each retained
direction while reducing the number of independently constructed
directions.
BiTE accounts for eight of the nine mean-based nondominated operating
points in each panel of Fig.~\ref{fig:parameter_accuracy}.

\subsection{Representation and Threshold Analysis}
\label{sec:representation_analysis}

We compare the effective rank of representations produced by BiTE and
uniform scalar collapse. For each modified BinaryViT layer, we arrange
the representations of identical held-out inputs into a matrix
$\mathbf{X}_{\ell}$, with feature dimensions as columns and sampled token positions
as rows. Given singular values $\{\sigma_i\}$, we compute effective rank as

\begin{equation}
    \operatorname{erank}(\mathbf{X}_{\ell})
    =
    \exp\!\left(
        -\sum_i p_i\log p_i
    \right),
    \qquad
    p_i=\frac{\sigma_i}{\sum_j\sigma_j}.
\label{eq:effective_rank}
\end{equation}

\begin{figure*}[t]
    \centering
    \includegraphics[width=0.47\textwidth]
    {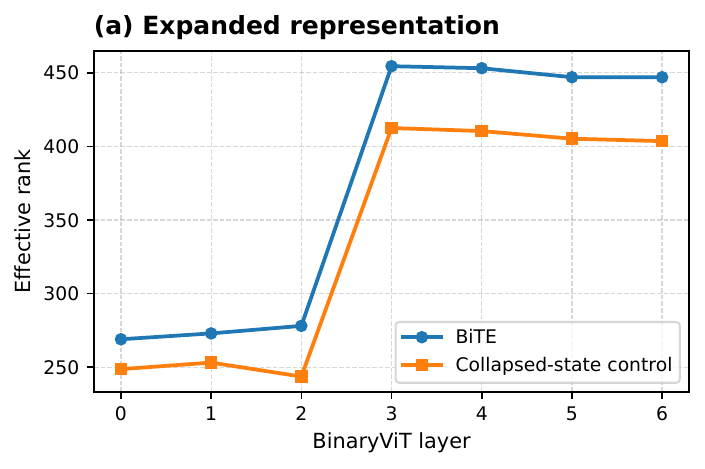}
    \hfill
    \includegraphics[width=0.47\textwidth]
    {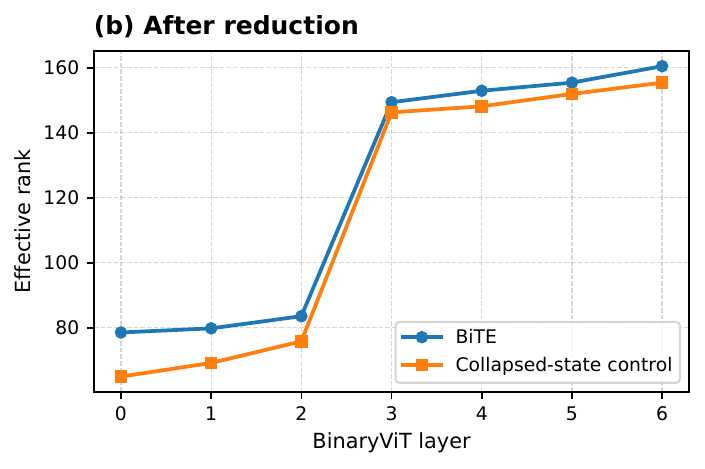}
    \caption{\textbf{Effective rank of BinaryViT representations.}
    Effective rank of BiTE and the collapsed-state control
    (a) immediately after threshold expansion and
    (b) after reduction to the same intermediate dimensionality. Curves show
    one matched seed-0 checkpoint pair evaluated on identical held-out inputs;
    they are not averages over runs.}
    \label{fig:effective_rank}
\end{figure*}
As shown in Fig.~\ref{fig:effective_rank}, BiTE has higher measured effective
rank than the collapsed-state control in every modified layer, both
immediately after threshold expansion and after reduction to the same
intermediate dimensionality. The difference ranges from $19.82$ to $43.38$
after expansion and from $3.13$ to $13.57$ after reduction. The expansion-level difference is partly structural because
the collapsed control repeats a scalar response across multiple feature
coordinates. Its persistence
after both representations are reduced to the same dimensionality is therefore
the more relevant observation. Nevertheless, this is a descriptive measurement
from one matched run and does not establish that the rank difference causes the
accuracy difference.

Across the three CIFAR-100 runs, the smallest final inter-threshold gap in any
modified layer exceeds $1.82$, compared with the initialized gap of $2.0$.
Thus, the thresholds remain separated at the end of training, although this
measurement alone does not demonstrate substantial threshold adaptation.

\begin{table}[!t]
\centering
\small
\setlength{\tabcolsep}{3.5pt}
\begin{tabular}{lccc}
\toprule
\textbf{Configuration} &
\textbf{Params} &
\textbf{Top-1 (\%)} &
\textbf{Top-5 (\%)} \\
\midrule
Single-sign baseline
    & $22.492$M
    & $59.952$
    & $82.464$ \\
BiTE, layers 0--6
    & $22.476$M
    & $59.392$
    & $82.240$ \\
BiTE, layers 0--6, $\mathrm{WD}_{t}=0$
    & $22.476$M
    & $59.488$
    & $82.600$ \\
BiTE, layers 0--2, $\mathrm{WD}_{t}=0$
    & $22.488$M
    & $\mathbf{60.016}$
    & $\mathbf{82.648}$ \\
\bottomrule
\end{tabular}
\caption{\textbf{Preliminary BinaryViT results on ImageNet.}
Each configuration is represented by one run. Top-1 and Top-5 are reported
from the checkpoint with the highest validation Top-1 accuracy.
$\mathrm{WD}_{t}=0$ denotes zero weight decay on threshold parameters.}
\label{tab:imagenet}
\end{table}
\subsection{Preliminary ImageNet Evaluation}
\label{sec:imagenet}

Applying BiTE selectively to layers 0--2 reaches $60.016\%$ Top-1 and
$82.648\%$ Top-5, compared with $59.952\%$ and $82.464\%$ for the
single-sign baseline, with nearly identical parameter counts. Because each
configuration is represented by one run, the small $0.064$-point Top-1
difference should be treated as descriptive rather than as evidence of a
reliable improvement.

Extending BiTE through layers 0--6 lowers Top-1 accuracy to $59.392\%$.
Removing threshold weight decay increases it to $59.488\%$, which remains
below the baseline. These results suggest that BiTE placement matters under
the evaluated ImageNet training configuration. Additional placement,
regularization, and occupancy experiments are provided in the supplementary
material.

\section{Conclusion}
\label{sec:conclusion}

We introduced \textbf{BiTE} (Branch-Preserving Binary Threshold Expansion),
which evaluates each preactivation direction using multiple ordered thresholds
and preserves the resulting binary responses as separate inputs to the next
learned operator. This allows the following operator to learn distinct effects
for different threshold crossings of the same feature direction. BiTE also
enables a controllable trade-off between the number of independently
constructed directions and the number of ordered responses extracted from
each direction.

Across BinaryViT experiments, BiTE improves mean final-epoch accuracy on
CIFAR-10 and CIFAR-100 with nearly unchanged trainable parameter counts.
The matched uniform-collapse control supports the value of preserving
threshold responses as separately transformable features relative to the
evaluated scalar collapse. Binary VGG-16 experiments further demonstrate a
favorable observed parameter--accuracy trade-off compared with conventional
width scaling, while the IE-Net experiment provides supporting evidence on a
convolutional architecture. The effective-rank analysis is also consistent
with BiTE retaining a less redundant representation after reduction.

A preliminary ImageNet study further shows that selective BiTE can be
integrated into the evaluated large-scale training configuration with nearly
unchanged model size. Overall, the results support branch-preserving threshold expansion
as an effective approach for improving binary representations and parameter
efficiency. Future work will evaluate BiTE on larger transformer backbones and
characterize its hardware-level efficiency in terms of operation count,
latency, memory traffic, and energy.

\setlength{\leftmargini}{20pt}
\makeatletter\def\@listi{\leftmargin\leftmargini \topsep .5em \parsep .5em \itemsep .5em}
\def\@listii{\leftmargin\leftmarginii \labelwidth\leftmarginii \advance\labelwidth-\labelsep \topsep .4em \parsep .4em \itemsep .4em}
\def\@listiii{\leftmargin\leftmarginiii \labelwidth\leftmarginiii \advance\labelwidth-\labelsep \topsep .4em \parsep .4em \itemsep .4em}\makeatother

\setcounter{secnumdepth}{0}
\renewcommand\thesubsection{\arabic{subsection}}
\renewcommand\labelenumi{\thesubsection.\arabic{enumi}}

\bibliography{aaai2026}

@article{hubara2016binarized,
  title={Binarized neural networks},
  author={Hubara, Itay and Courbariaux, Matthieu and Soudry, Daniel and El-Yaniv, Ran and Bengio, Yoshua},
  journal={Advances in neural information processing systems},
  volume={29},
  year={2016}
}

@inproceedings{rastegari2016xnor,
  title={Xnor-net: Imagenet classification using binary convolutional neural networks},
  author={Rastegari, Mohammad and Ordonez, Vicente and Redmon, Joseph and Farhadi, Ali},
  booktitle={European conference on computer vision},
  pages={525--542},
  year={2016},
  organization={Springer}
}

@inproceedings{liu2018bi,
  title={Bi-real net: Enhancing the performance of 1-bit cnns with improved representational capability and advanced training algorithm},
  author={Liu, Zechun and Wu, Baoyuan and Luo, Wenhan and Yang, Xin and Liu, Wei and Cheng, Kwang-Ting},
  booktitle={Proceedings of the European conference on computer vision (ECCV)},
  pages={722--737},
  year={2018}
}

@inproceedings{qin2020forward,
  title={Forward and backward information retention for accurate binary neural networks},
  author={Qin, Haotong and Gong, Ruihao and Liu, Xianglong and Shen, Mingzhu and Wei, Ziran and Yu, Fengwei and Song, Jingkuan},
  booktitle={Proceedings of the IEEE/CVF conference on computer vision and pattern recognition},
  pages={2250--2259},
  year={2020}
}

@inproceedings{xu2021recu,
  title={Recu: Reviving the dead weights in binary neural networks},
  author={Xu, Zihan and Lin, Mingbao and Liu, Jianzhuang and Chen, Jie and Shao, Ling and Gao, Yue and Tian, Yonghong and Ji, Rongrong},
  booktitle={Proceedings of the IEEE/CVF international conference on computer vision},
  pages={5198--5208},
  year={2021}
}

@Article{ding2022ienet,
AUTHOR = {Ding, Rui and Liu, Haijun and Zhou, Xichuan},
TITLE = {IE-Net: Information-Enhanced Binary Neural Networks for Accurate Classification},
JOURNAL = {Electronics},
VOLUME = {11},
YEAR = {2022},
NUMBER = {6},
ARTICLE-NUMBER = {937},
}

@inproceedings{liu2020reactnet,
  title={Reactnet: Towards precise binary neural network with generalized activation functions},
  author={Liu, Zechun and Shen, Zhiqiang and Savvides, Marios and Cheng, Kwang-Ting},
  booktitle={European conference on computer vision},
  pages={143--159},
  year={2020},
  organization={Springer}
}

@inproceedings{zhang2022dynamic,
  title={Dynamic binary neural network by learning channel-wise thresholds},
  author={Zhang, Jiehua and Su, Zhuo and Feng, Yanghe and Lu, Xin and Pietik{\"a}inen, Matti and Liu, Li},
  booktitle={ICASSP 2022-2022 IEEE International Conference on Acoustics, Speech and Signal Processing (ICASSP)},
  pages={1885--1889},
  year={2022},
  organization={IEEE}
}

@inproceedings{lee2023insta,
  title={Insta-bnn: Binary neural network with instance-aware threshold},
  author={Lee, Changhun and Kim, Hyungjun and Park, Eunhyeok and Kim, Jae-Joon},
  booktitle={Proceedings of the IEEE/CVF International Conference on Computer Vision},
  pages={17325--17334},
  year={2023}
}

@inproceedings{tu2022adabin,
  title={Adabin: Improving binary neural networks with adaptive binary sets},
  author={Tu, Zhijun and Chen, Xinghao and Ren, Pengju and Wang, Yunhe},
  booktitle={European conference on computer vision},
  pages={379--395},
  year={2022},
  organization={Springer}
}

@inproceedings{falkena2023lab,
  title={LAB: Learnable activation binarizer for binary neural networks},
  author={Falkena, Sieger and Jamali-Rad, Hadi and van Gemert, Jan},
  booktitle={Proceedings of the IEEE/CVF Winter Conference on Applications of Computer Vision},
  pages={6425--6434},
  year={2023}
}

@article{monroy2024designed,
  title={Designed dithering Sign activation for binary neural networks},
  author={Monroy, Brayan and Estupinan, Juan and Gelvez-Barrera, Tatiana and Bacca, Jorge and Arguello, Henry},
  journal={IEEE Journal of Selected Topics in Signal Processing},
  volume={18},
  number={6},
  pages={1100--1107},
  year={2024},
  publisher={IEEE}
}

@article{lin2017towards,
  title={Towards accurate binary convolutional neural network},
  author={Lin, Xiaofan and Zhao, Cong and Pan, Wei},
  journal={Advances in neural information processing systems},
  volume={30},
  year={2017}
}

@inproceedings{ghasemzadeh2018rebnet,
  title={ReBNet: Residual binarized neural network},
  author={Ghasemzadeh, Mohammad and Samragh, Mohammad and Koushanfar, Farinaz},
  booktitle={2018 IEEE 26th annual international symposium on field-programmable custom computing machines (FCCM)},
  pages={57--64},
  year={2018},
  organization={IEEE}
}

@inproceedings{zhuang2019structured,
  title={Structured binary neural networks for accurate image classification and semantic segmentation},
  author={Zhuang, Bohan and Shen, Chunhua and Tan, Mingkui and Liu, Lingqiao and Reid, Ian},
  booktitle={Proceedings of the IEEE/CVF conference on computer vision and pattern recognition},
  pages={413--422},
  year={2019}
}

@article{zhou2016dorefa,
  title={Dorefa-net: Training low bitwidth convolutional neural networks with low bitwidth gradients},
  author={Zhou, Shuchang and Wu, Yuxin and Ni, Zekun and Zhou, Xinyu and Wen, He and Zou, Yuheng},
  journal={arXiv preprint arXiv:1606.06160},
  year={2016}
}

@article{choi2018pact,
  title={Pact: Parameterized clipping activation for quantized neural networks},
  author={Choi, Jungwook and Wang, Zhuo and Venkataramani, Swagath and Chuang, Pierce I-Jen and Srinivasan, Vijayalakshmi and Gopalakrishnan, Kailash},
  journal={arXiv preprint arXiv:1805.06085},
  year={2018}
}

@article{esser2019learned,
  title={Learned step size quantization},
  author={Esser, Steven K and McKinstry, Jeffrey L and Bablani, Deepika and Appuswamy, Rathinakumar and Modha, Dharmendra S},
  journal={arXiv preprint arXiv:1902.08153},
  year={2019}
}

@article{courbariaux2015binaryconnect,
  title={Binaryconnect: Training deep neural networks with binary weights during propagations},
  author={Courbariaux, Matthieu and Bengio, Yoshua and David, Jean-Pierre},
  journal={Advances in neural information processing systems},
  volume={28},
  year={2015}
}

@article{shi2025expanding,
  title={Expanding-and-shrinking binary neural networks},
  author={Shi, Xulong and Sun, Caiyi and Qi, Zhi and Hao, Liu and Yang, Xiaodong},
  journal={arXiv preprint arXiv:2503.23709},
  year={2025}
}

@inproceedings{le2023binaryvit,
  title={Binaryvit: Pushing binary vision transformers towards convolutional models},
  author={Le, Phuoc-Hoan Charles and Li, Xinlin},
  booktitle={Proceedings of the IEEE/CVF conference on computer vision and pattern recognition},
  pages={4665--4674},
  year={2023}
}

@article{xiao2024binaryvit,
  title={BinaryViT: Toward efficient and accurate binary vision transformers},
  author={Xiao, Junrui and Li, Zhikai and Li, Jianquan and Yang, Lianwei and Gu, Qingyi},
  journal={IEEE Transactions on Circuits and Systems for Video Technology},
  volume={35},
  number={1},
  pages={195--206},
  year={2024},
  publisher={IEEE}
}

@inproceedings{li2024bi,
  title={Bi-vit: Pushing the limit of vision transformer quantization},
  author={Li, Yanjing and Xu, Sheng and Lin, Mingbao and Cao, Xianbin and Liu, Chuanjian and Sun, Xiao and Zhang, Baochang},
  booktitle={Proceedings of the AAAI conference on artificial intelligence},
  volume={38},
  number={4},
  pages={3243--3251},
  year={2024}
}

@inproceedings{wang2024bvt,
  title={Bvt-ima: Binary vision transformer with information-modified attention},
  author={Wang, Zhenyu and Luo, Hao and Xie, Xuemei and Wang, Fan and Shi, Guangming},
  booktitle={Proceedings of the AAAI Conference on Artificial Intelligence},
  volume={38},
  number={14},
  pages={15761--15769},
  year={2024}
}

@inproceedings{gao2025bhvit,
  title={Bhvit: Binarized hybrid vision transformer},
  author={Gao, Tian and Zhang, Yu and Zhang, Zhiyuan and Liu, Huajun and Yin, Kaijie and Xu, Chengzhong and Kong, Hui},
  booktitle={Proceedings of the Computer Vision and Pattern Recognition Conference},
  pages={3563--3572},
  year={2025}
}

@article{krizhevsky2009learning,
  title={Learning multiple layers of features from tiny images},
  author={Krizhevsky, Alex and Hinton, Geoffrey and others},
  year={2009},
  publisher={Toronto, ON, Canada}
}

@inproceedings{deng2009imagenet,
  title={Imagenet: A large-scale hierarchical image database},
  author={Deng, Jia and Dong, Wei and Socher, Richard and Li, Li-Jia and Li, Kai and Fei-Fei, Li},
  booktitle={2009 IEEE conference on computer vision and pattern recognition},
  pages={248--255},
  year={2009},
  organization={Ieee}
}

@inproceedings{gupta2025ab,
  title={AB-STE: Adaptive Blended Gradient Estimation for Efficient Binarized Networks},
  author={Gupta, Siddharth and Kumar, Akash},
  booktitle={Joint European Conference on Machine Learning and Knowledge Discovery in Databases},
  pages={125--140},
  year={2025},
  organization={Springer}
}

\makeatletter
\@ifundefined{isChecklistMainFile}{
  \newif\ifreproStandalone
  \reproStandalonetrue
}{
  \newif\ifreproStandalone
  \reproStandalonefalse
}
\makeatother

\ifreproStandalone

\setlength{\pdfpagewidth}{8.5in}
\setlength{\pdfpageheight}{11in}

\frenchspacing

\fi
\setlength{\leftmargini}{20pt}
\makeatletter\def\@listi{\leftmargin\leftmargini \topsep .5em \parsep .5em \itemsep .5em}
\def\@listii{\leftmargin\leftmarginii \labelwidth\leftmarginii \advance\labelwidth-\labelsep \topsep .4em \parsep .4em \itemsep .4em}
\def\@listiii{\leftmargin\leftmarginiii \labelwidth\leftmarginiii \advance\labelwidth-\labelsep \topsep .4em \parsep .4em \itemsep .4em}\makeatother

\setcounter{secnumdepth}{0}
\renewcommand\thesubsection{\arabic{subsection}}
\renewcommand\labelenumi{\thesubsection.\arabic{enumi}}

\end{document}